\RequirePackage{fix-cm}
\documentclass[twocolumn]{svjour3}          % twocolumn
\smartqed  % flush right qed marks, e.g. at end of proof
\usepackage{array}
\usepackage{graphicx}
\usepackage{amsmath}
\usepackage{amssymb}
\usepackage{amsfonts}
\usepackage{multirow}
\usepackage{rotating}
\usepackage{epstopdf}
\usepackage{tabu}
\usepackage{subfigure}
\usepackage{makecell}
\begin{document}

\title{An Efficient Deep Learning Approach Using Improved Generative Adversarial Networks for Incomplete Information Completion of Self-driving Vehicles%\thanks{Grants or other notes
}


\author{Jingzhi Tu \and
	Gang Mei \and
	Francesco Piccialli %etc.
}


\institute{J. Tu \and G. Mei \at
	School of Engineering and Technolgy, \\China University of Geosciences, Beijing, China \\
	\email{gang.mei@cugb.edu.cn}
	\and
	F. Piccialli \at
	Department of Mathematics and Applications ``R. Caccioppoli", \\
	University of Naples Federico II, Italy \\
}

\date{Received: date / Accepted: date}

\maketitle

\begin{abstract}
Autonomous driving is the key technology of intelligent logistics in Industrial Internet of Things (IIoT). In autonomous driving, the appearance of incomplete point clouds losing geometric and semantic information is inevitable owing to limitations of occlusion, sensor resolution, and viewing angle when the Light Detection And Ranging (LiDAR) is applied. The emergence of incomplete point clouds, especially incomplete vehicle point clouds, would lead to the reduction of the accuracy of autonomous driving vehicles in object detection, traffic alert, and collision avoidance. Existing point cloud completion networks, such as Point Fractal Network (PF-Net), focus on the accuracy of point cloud completion, without considering the efficiency of inference process, which makes it difficult for them to be deployed for vehicle point cloud repair in autonomous driving. To address the above problem, in this paper, we propose an efficient deep learning approach to repair incomplete vehicle point cloud accurately and efficiently in autonomous driving. In the proposed method, an efficient downsampling algorithm combining incremental sampling and one-time sampling is presented to improves the inference speed of the PF-Net based on Generative Adversarial Network (GAN). To evaluate the performance of the proposed method, a real dataset is used, and an autonomous driving scene is created, where three incomplete vehicle point clouds with 5 different sizes are set for three autonomous driving situations. The improved PF-Net can achieve the speedups of over 19x with almost the same accuracy when compared to the original PF-Net. Experimental results demonstrate that the improved PF-Net can be applied to efficiently complete vehicle point clouds in autonomous driving.
%\subclass{MSC code1 \and MSC code2 \and more}
\end{abstract}

\section{Introduction}
\label{intro}
With the wide use of internet-based sensors, a new concept termed as the Internet of Things (IoT) has emerged to facilitate the development of modern industry \cite{1,19}. The ingenious application of intelligent logistics by introducing Industrial Internet of Things (IIoT) significantly promotes the development of the transportation industry, where autonomous driving is the key technology. 3D vision using LiDAR has widely used in autonomous driving due to its lower cost, more robust, richer information, and meeting the mass-production standards \cite{2}. Typically, to detect obstacles and other relevant driving information, the computer vision system receives and analyzes the original point cloud from LiDAR, where the appearance of incomplete point clouds is inevitable owing to limitations of weather, sensor capability, and viewing angle \cite{3}. The emergence of incomplete point clouds, especially fragmentary vehicle point clouds, may cause the reduction of the accuracy of autonomous driving vehicles in object detection, traffic alert, and collision avoidance.

For the above problem, traditional methods, such as interpolation algorithms \cite{20,21,22}, cannot describe the complex geometric characteristics of the vehicle without reference points. In recent years, deep learning is gradually widely applied to 3D vision systems with the rapid development of machine learning. Since the advent of the classic PointNet \cite{4} showing powerful point clouds operating ability, various deep-learning based approaches are developed to solve 3D vision tasks directly \cite{5,6,7,8}. For the point cloud completion, Panos et al. \cite{1} proposed the first deep-learning based network by utilizing an Encoder-Decoder framework. Point Completion Network (PCN) \cite{10} performs shape completion tasks by combining the advantages of Latent-space Generative Adversarial Network (L-GAN) \cite{9} and FoldingNet \cite{11}. Moreover, the extensions of Convolutional Neural Network (CNN) in 3D have been demonstrated to perform well when 3D voxel grids are used to represent 3D shapes \cite{12,13,14}, where the CNN is computationally expensive for high voxel resolution. Muhammad et al. \cite{15} introduced the first reinforcement learning agent controlled Generative Adversarial Network (GAN) to generate point clouds quickly. Moreover, to repair the incomplete point cloud without modifying the position of known points, an unsupervised point repair network based on GAN, Point Fractal Network (PF-Net) \cite{3}, is proposed by paying attention to the local details of an object, which makes it perform well in accuracy. 

However, these existing deep-learning based networks for point cloud completion focus on the accuracy of point cloud completion, without considering the efficiency of the inference process, which makes it difficult for them to be deployed for vehicle point cloud repair in autonomous driving, e.g., the state-of-the-art point cloud repair network, PF-Net, has unsatisfactory inference speed due to the low efficiency of its downsampling module. 

Downsampling is a basic operation for large-scale point clouds, which can improve the efficiency of subsequent algorithms, including 3D reconstruction \cite{28,29}, point cloud generation \cite{30,31}, and recognition \cite{32,33}. Uniform downsampling, one of downsampling, is used to obtain uniform sparse point clouds consistent with the geometric shape of the target, so as to remove noise while preserving the geometric features of the target. In the uniform downsampling algorithms, there are two commonly used algorithms: Iterative Farthest Point Sampling (IFPS) \cite{17} and cell sampling \cite{34}. As an incremental sampling algorithm \cite{35,36}, the IFPS is widely applied in point cloud networks such as PointNet++ \cite{17} and PF-Net because of its capability to control number of sampling points. However, the IFPS has poor performance in efficiency due to the characteristics of the incremental sampling algorithm. Cell sampling, as a one-time sampling method \cite{36}, has an excellent performance in efficiency, but it cannot precisely control the number of sampling points.

In this paper, our objective is to repair incomplete vehicle point clouds accurately and efficiently, which can serve object detection, traffic alert, and collision avoidance to improve driving safety in autonomous driving. To address the problem, we proposed to complete incomplete vehicle point clouds of self-driving vehicles using the deep learning approach, i.e., the improved PF-Net. In the improved method, we proposed an efficient uniform downsampling algorithm (Cell-IFPS) by combining the advantages of the IFPS and cell sampling, which significantly improves the inference efficiency of original PF-Net.

The contributions of this paper can be summarized as follows:

(1) We employ a deep learning approach (i.e., the improved PF-Net) to repair incomplete vehicle point clouds that cannot be addressed by traditional methods in autonomous driving.

(2) We improve the efficiency of original PF-Net to meet the requirement of point cloud completion of self-driving vehicles, where an efficient uniform downsampling algorithm by combining the advantages of the IFPS and cell sampling (i.e., the high efficiency and the ability to precisely control the number of sampling points) is presented.

(3) We evaluate the method on a real-world vehicle point cloud collected from a real vehicle self-driving process and three incomplete vehicle point clouds with 5 different sizes collected from high-fidelity vehicle models.

The rest of the paper is organized as follows. In Section \ref{Method}, the significance and the implementation process of incomplete vehicle information completion, and the architecture of the improved PF-Net are introduced. In Section \ref{Results}, the experimental design and data are briefly introduced, then the repair results of incomplete vehicle point clouds are shown. In Section \ref{dis}, we had some discussions based on the experimental results. In Section \ref{con}, some conclusions are obtained.

\section{Method}
\label{Method}

\subsection{Overview}
\begin{figure*}[!h]
	\centering
	\includegraphics[width=\textwidth]{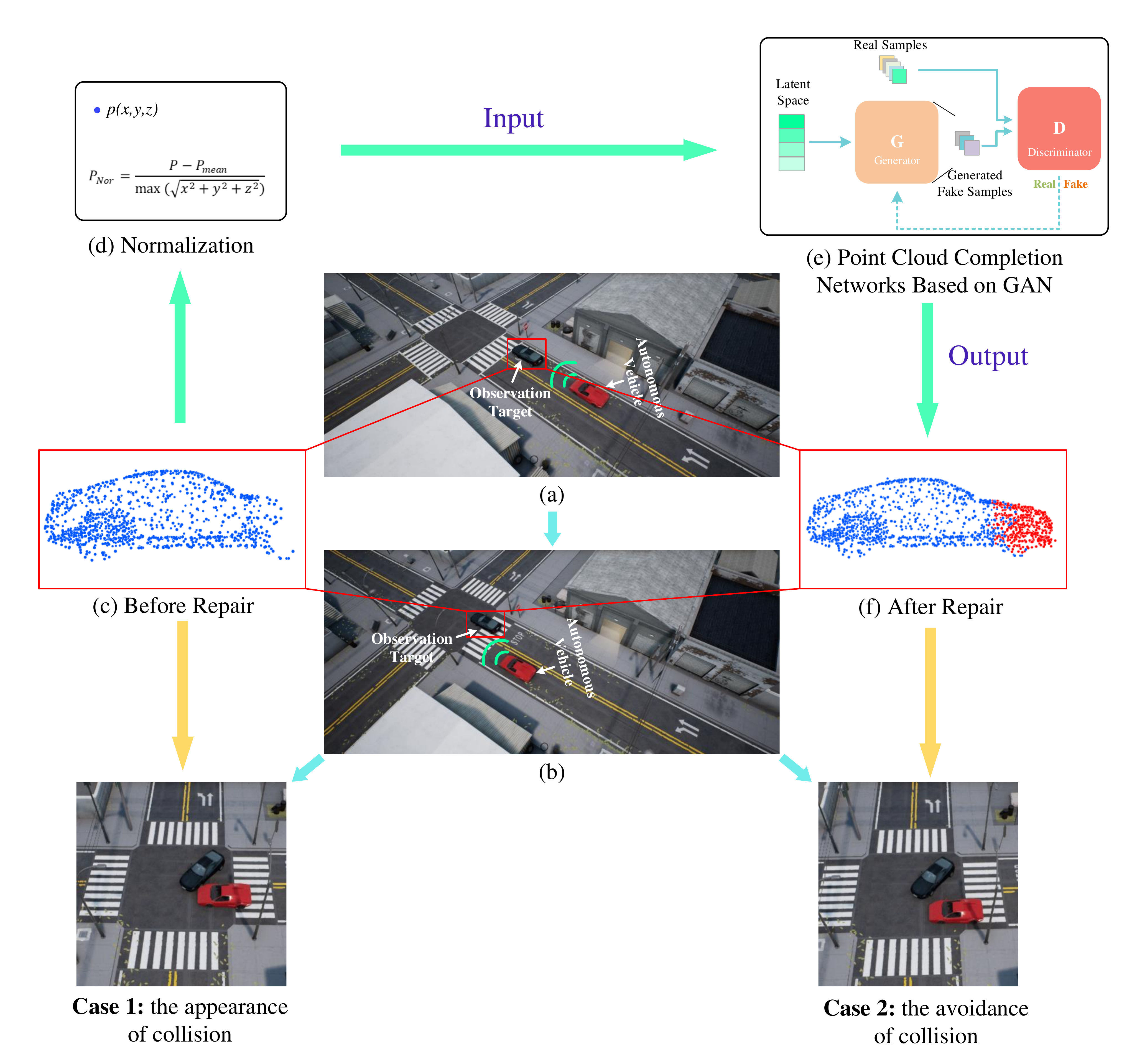}
	\caption{Illustration of the proposed deep learning approach using improved GAN for incomplete information completion of self-driving vehicle}
	\label{fig1}       % Give a unique label
\end{figure*}

In this paper, we propose to complete incomplete vehicle point clouds of self-driving vehicles using the deep learning approach due to the limitations of traditional data completion methods, such as interpolation algorithm. As demonstrated in Fig. \ref{fig1}, the incomplete point cloud obtained by using LiDAR is inevitable owing to limitations of viewing angle (see Fig. \ref{fig1}(a)-(c)), which may cause the self-driving vehicle to misjudge the size of the surrounding car, resulting in the appearance of collision (see Fig. \ref{fig1} Case 1). To avoid this potentially dangerous situation, we first normalize the incomplete vehicle point cloud obtained by using LiDAR and then input it into the point cloud completion networks based on GAN (i.e., the improved PF-Net) to repair the point incomplete cloud. Finally, the self-driving vehicle design a safe route according to the complete vehicle point cloud to avoid collision (see Fig. \ref{fig1} Case 2).

\subsection{Improved GAN-based Point Cloud Completion}

\begin{figure*}[!htb]
	\centering
	\includegraphics[width=\textwidth]{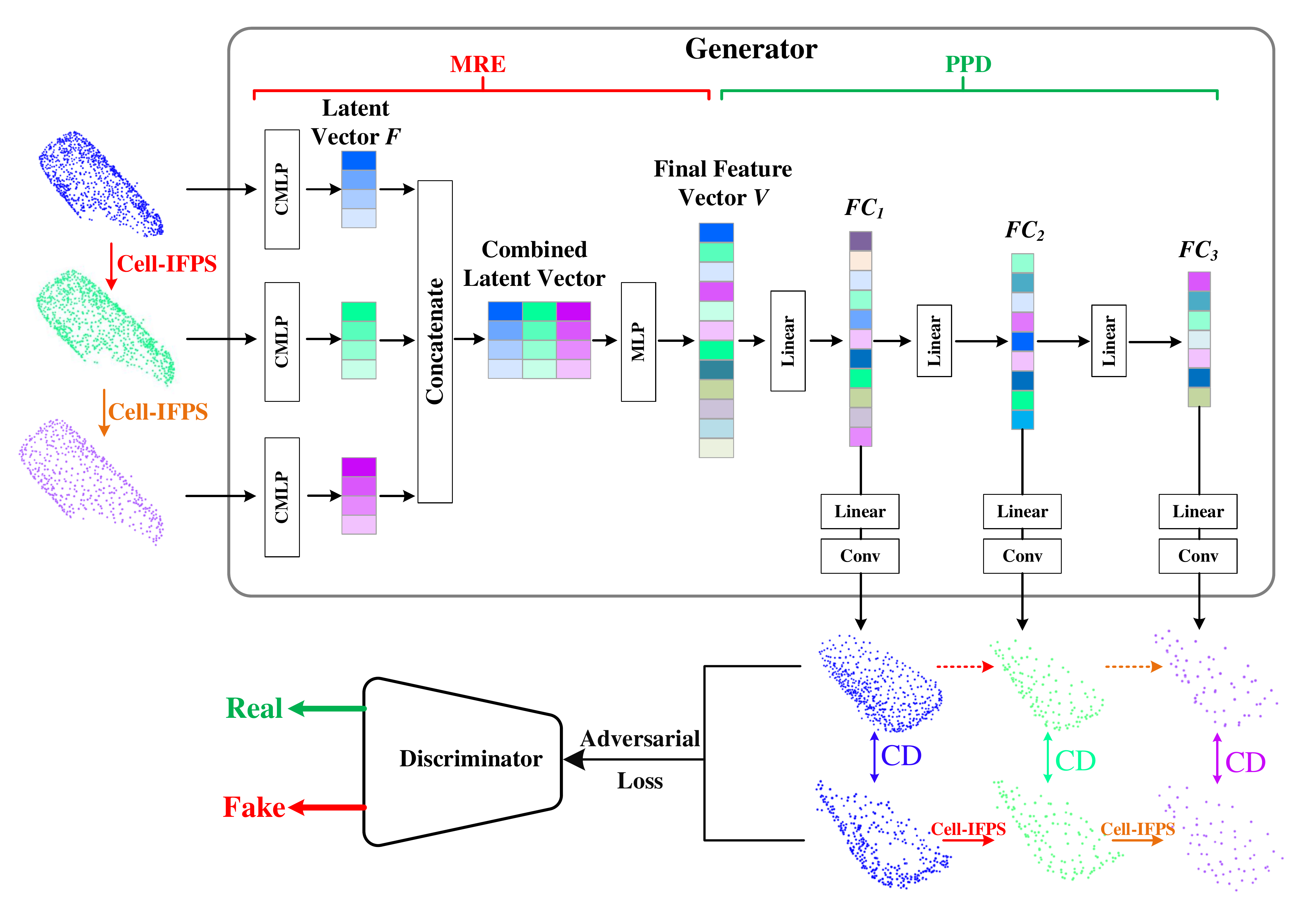}
	\caption{The overall architecture of improved PF-Net}
	\label{fig2}       % Give a unique label
\end{figure*}

In this section, the specific framework of improved PF-Net will be introduced (see Fig. \ref{fig2}), which predicts the missing part of the point cloud from its incomplete known configuration. Compared with the original PF-Net, we mainly modified the sampling step in the overall architecture. The overall architecture of improved PF-Net consists of three fundamental parts, including Multi-Resolution Encoder (MRE), Point Pyramid Decoder (PPD), and Discriminator Network. The PF-Net is a generative adversarial network \cite{16}, and the Generator consists of MRE and PPD.

First, the improved PF-Net uses Cell-IFPS to extract feature points from the input point cloud in the MRE, which are representative points that can map the whole shape of the input. Then, three scales point clouds including the extracted feature points and the original points are mapped into three individual latent vectors $F_i$ ($i=1,2,3$) by three independent Combined Multi-Layer Perceptions (CMLPs). Finally, the combined latent vector is mapped into the final feature vector $V$ by a Multi-Layer Perception (MLP) after those three individual latent vectors $F_i$ are concatenated (see Fig.~\ref{fig2}). In the improved PF-Net, CMLP is proposed as the feature extractor. The common MLP, such as PointNet-MLP, cannot combine the low-level and middle-level features well when it only applies the final single layer to obtain the latent vector. Compared with common MLP, the CMLP is found to be able to extract combined latent vector after concatenating multiple dimensional potential vectors stemed from the last four layers (low-level, middle-level, and high-level features), which leads to better use of multi-layer features.

The objective of PDD is to output point clouds with three different resolutions which represent the shape of the missing region by the final feature vector $V$ obtained in the previous step. Therefore, the vector $V$ is first transformed into three feature results ${FC}_i$ with different dimensions (where ${FC}_i=1024,512,256$ neurons, and $i=1,2,3$) by three fully-connected layers. Then, point clouds with different resolutions are predicted according to different features ${FC}_i$.

The loss function is created by combining multi-stage completion loss (MCS) and adversarial loss in the improved PF-Net. The MCS is used to minimize the difference between the predicted three stages of point cloud and Ground Truth (GT). The combined loss for three stages is as follows:
\begin{equation}
\begin{split}
L_{com}=&d_{CD1}\left(Y_{detail},Y_{GT}\right)+{\alpha d}_{CD2}\left(Y_{primary},Y^{'}_{GT}\right) \\
&+2{\alpha d}_{CD3}(Y_{secondary},Y^{''}_{GT})
\end{split}
\label{eq1}
\end{equation}
where $Y_{primary}$ is the primary center points which be predicted by ${FC}_3$, $Y_{secondary}$ is the secondary center points which be predicted by $Y_{primary}$ and ${FC}_2$. The generation of $Y_{detail}$ similar to $Y_{secondary}$. Moreover, $Y_{GT}$ is the ground truth of missing point cloud, and $Y^{'}_{GT}$ and $Y^{''}_{GT}$ are obtained from $Y_{GT}\ $by applying IFPS. $\alpha $ is the hyperparameter.

The equation of Chamber Distance (CD) is as follows:
\begin{equation}
d_{CD}(S_1,S_2)=\frac{1}{S_1}\sum_{x\in S_1}{{\mathop{\mathrm{min}}_{y\in S_2} {\big\|x-y\big\|}^2_2+\frac{1}{S_2}\sum_{y\in S_2}{{\mathop{\mathrm{min}}_{x\in S_1} {\big\|y-x\big\|}^2_2\ }}\ }}
\label{eq2}
\end{equation}
where $S_1$ is the predicted point cloud, and $S_2$ is the ground truth point cloud.

The adversarial loss which is inspired by GAN is defined as follows:
\begin{equation}
L_{adv}=\sum_{1\le i\le S}{{\mathrm{log} \left(D\left(y_i\right)\right)+\sum_{1\le j\le S}{{\mathrm{log} \left(1-D\left(F(x_i)\right)\right)\ }}\ }}
\label{eq3}
\end{equation}
where $x_i\in X$ which is the partial input, $y_i\in Y$ which is the real missing region, $i=1,2,\cdots ,S$ ($S$ is the dataset size of $X$,$\ Y$). $D$ is the discriminator which is used to distinguish the predicted missing point cloud and the real missing point cloud. The function $F$ is defined as $F(){{:=}}PPD(MRE())$ which maps the partial input $X$ into predicted missing region $Y^{'}$.

Finally, the joint loss of the MCS and the adversarial loss is defined as:
\begin{equation}
L={\lambda }_{c}L_{com}+{\lambda }_{a}L_{adv}
\label{eq4}
\end{equation}
where ${\lambda }_{c}$ is the weight of completion loss $L_{com}$, ${\lambda }_{a}$ is the weight of adversarial loss $L_{adv}$, ${\lambda }_{c}+{\lambda }_{a}=1$.

\subsection{Cell-IFPS}

Herein, the proposed Cell-IFPS is introduced in detail because it is the key to improve the efficiency of the original PF-Net. 

\begin{figure*}[!htb]
	\centering
	\includegraphics[width=\textwidth]{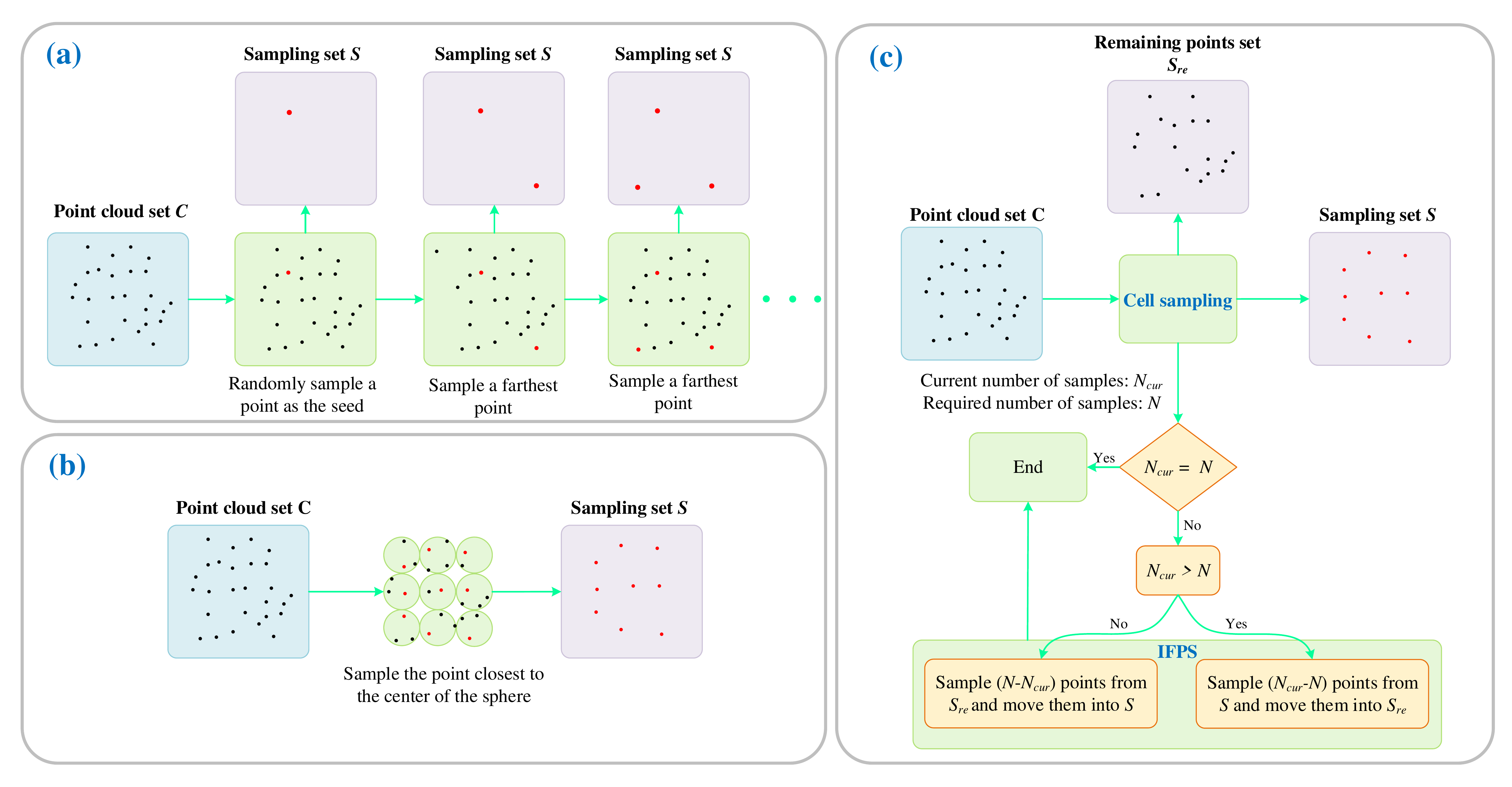}
	\caption{An illustration  of downsampling algorithms ((a) IFPS; (b) Cell sampling; (c) Cell-IFPS.)}
	\label{fig3}       % Give a unique label
\end{figure*}

In the original PF-Net, the inefficient IFPS is used due to its ability to precisely control the number of samples, resulting in the original PF-Net inefficiencies for autonomous driving scenarios. As shown in Fig. \ref{fig3}(a), the IFPS uses a circular iterative approach to sampling, the characteristics of which make it inefficient and difficult to accelerate using parallel strategies \cite{23,24,25}. In the IFPS, we first establish the point cloud set $C$ and the sampling set $S$ where the sampling points are stored, and then randomly sample a point from $C$ as the seed (the starting sampling point) and puts it into set $S$. Finally, we sample one farthest point at a time and puts it into set $S$ until the number of sampling points meets the requirements. The farthest point is the one with the greatest distance from the set $S$ among the remaining points set $S_{re}$, and the greatest distance is defined as follows:
\begin{equation}
d_{max}\mathrm{=}{\mathop{\mathrm{max}}_{x_i\in S_{re}} ({min (dist(x_i,S))\ })\ }
\label{eq5}
\end{equation}
where $dist(x_i,S)$ indicates that the distance between $x_i$ and each point in the set $S$ is calculated.

As shown in Fig. \ref{fig3}(b), there is another common method: cell sampling, which downsamples the point cloud by constructing spheres with a specified radius, and outputs the point in each sphere closest to the center of the sphere as the sampling point. The sampling method is extremely efficient, but the number of samples can only be controlled by adjusting the radius of the sphere.

Therefore, to sample efficiently and exactly control the number of samples, the efficient Cell-IFPS by combining the advantages of the IFPS and cell sampling is proposed, as shown in Fig. \ref{fig3}(c), the sampling steps are as follows:

(1) Set number of samples $N$.

(2) Sample the point cloud using the cell sampling, and create the point cloud $C$, the sampling set $S$, and the remaining points set $S_{re}$.

(3) Determine whether the current number of samples $N_{cur}$ in the set $S$. is greater than the number of samples $N$. If it is equal, the sampling is finished, if not, the next step is performed.

(4) Randomly sample a point from $S$ as the seed, and determine whether the current number of samples $N_{cur}$ in the set $S$ is equal to the number of samples $N$. If $N_{cur}>N$, the IFPS is used to sample $(N_{cur}-N$) points from $S$ and these points are moved to $S_{re}$; If $N_{cur}<N$, the IFPS is used to sample $(N-N_{cur}$) points from $S_{re}$ and these points are moved to $S$.

\section{Results}
\label{Results}
In this section, to evaluate the performance of the improved PF-Net, we use a real-world vehicle point cloud and three incomplete vehicle point clouds collected from high-fidelity models to test it. 

\subsection{Experimental Environment}
The details of the experimental environment are listed in Table \ref{tab1}.
\begin{table}[!ht]
	\centering
	\caption{The platforms used for testing}
	\label{tab1}
 \begin{tabular}{p{1.1in}p{1.4in}} \hline\noalign{\smallskip}
    \textbf{Specifications}	& \textbf{Details} \\ \noalign{\smallskip}\hline\noalign{\smallskip}
	\textbf{CPU} & Intel Xeon 5118  \\ 
	\textbf{CPU RAM} & 128 GB \\ 
	\textbf{CPU Frequency} & 2.30 GHz \\ 
	\textbf{GPU} & Quadro P6000 \\ 
	\textbf{CUDA Cores} & 3840 \\ 
	\textbf{GPU RAM} & 24 GB \\ 
	\textbf{OS} & Window 10 \\ 
	\textbf{Python} & Version 3.7 \\ 
	\textbf{Pytorch} & Version 1.8 \\ 
	\textbf{CUDA version} & Version 10.2 \\ \noalign{\smallskip}\hline
 \end{tabular}
\end{table}

\subsection{Data Description}

The training dataset of the model is the ShapeNet Part \cite{39}, which is a dataset that selects 16 categories from the ShapeNetCore dataset \cite{40} and annotated with semantic information for the semantic segmentation task of point clouds.

\begin{figure*}[!htb]
	\centering
	\includegraphics[width=\textwidth]{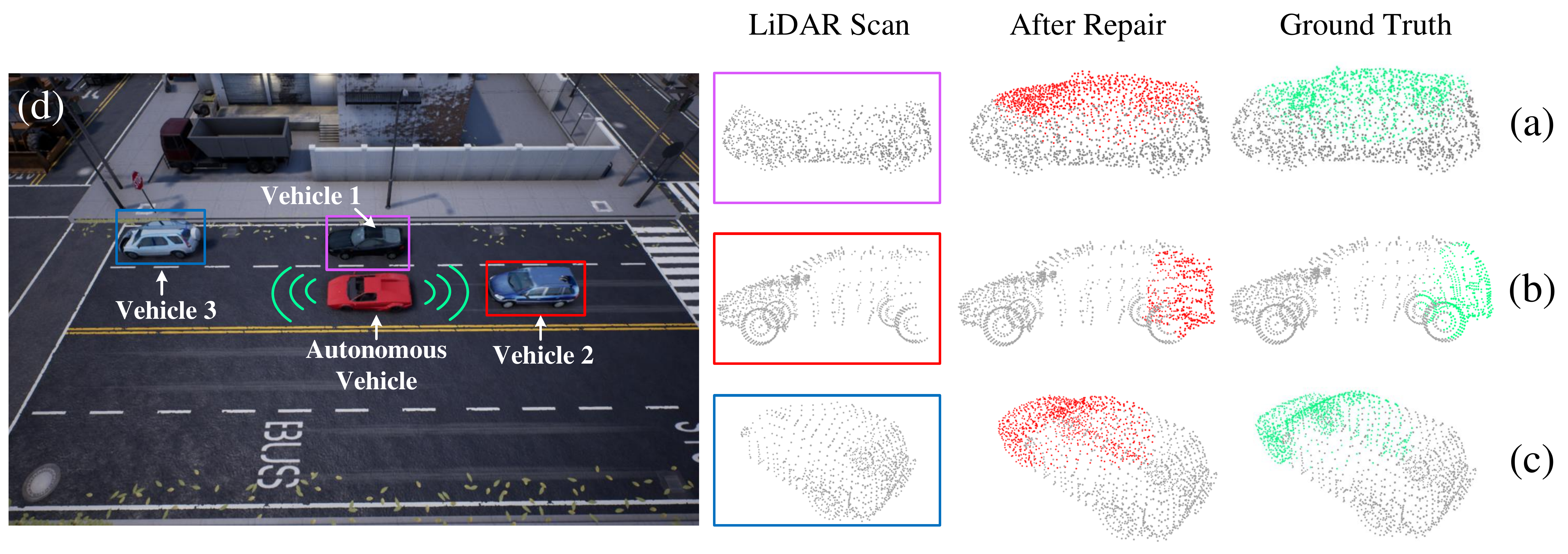}
	\caption{An Illustration of three incomplete vehicle point clouds completion for the experimental test in autonomous driving}
	\label{fig4}       % Give a unique label
\end{figure*}

As shown in Fig. \ref{fig4}, to simulate real autonomous driving scenarios, we set up a two-way road with two lanes in an industrial estate (see Fig. \ref{fig3}(d)). In our experiment, there are three different cars around the red autonomous vehicle, and in addition, we obtained point clouds of 5 different scales from each car. The missing regions of the vehicle point clouds obtained by LiDAR scanning are also different due to the limitation of viewing angle (LiDAR Scan in Fig. \ref{fig4}). Therefore, we have completed these three missing vehicle point clouds that are common in the real world (see Fig. \ref{fig4}(a)-(c)). It should be noted that the vehicle point clouds with missing regions are actually complete, but for testing, incomplete vehicle point clouds are assumed.

\begin{figure}[!htb]
	%\centering
	\includegraphics[width=0.45\textwidth]{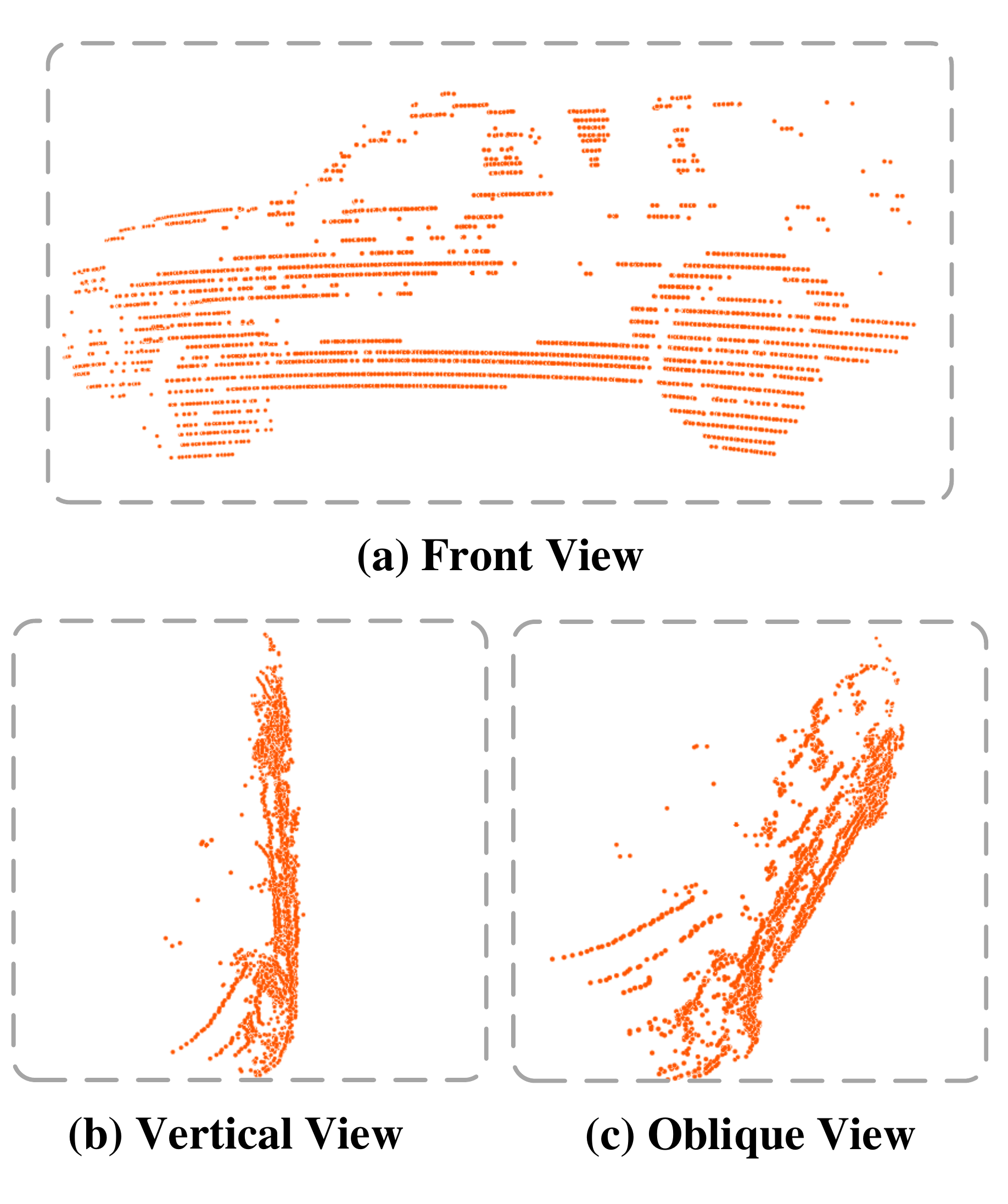}
	\caption{An incomplete vehicle point cloud from the real self-driving dataset KITTIT}
	\label{fig5}       % Give a unique label
\end{figure}

As shown in Fig. \ref{fig5}, there is an incomplete vehicle point cloud with 2341 points from the famous KITTIT dataset \cite{37,38}, which is the real dataset collected in the self-driving scenarios. Thus, it should be noted that the real vehicle point cloud is incomplete, and we don't have its real missing region.

\subsection{Completion Results on Vehicle Point Clouds of Different Scales}
In this section, we used the original PF-Net and the improved PF-Net to complete the vehicle point clouds of different sizes, where we utilized the original PF-Net as a baseline. Then, we used Pred $\mathrm{\to}$ GT (prediction to ground truth) error and GT $\mathrm{\to}$ Pred (ground truth to prediction) error to evaluate the results of completion. Finally, we obtained the running time of each step of the models.

\subsubsection{Completion Accuracy on Vehicle Point Clouds of Different Scales}

To evaluate the accuracy of completion results in autonomous driving, we use Pred $\mathrm{\to}$ GT error and GT $\mathrm{\to}$ Pred error as evaluation metrics, which are widely used to evaluate the quality of point cloud generation \cite{10,17}. Pred $\mathrm{\to}$ GT error means the mean squared distance between each predicted point and its nearest point in the ground truth. It measures the difference between the prediction and the truth. GT $\mathrm{\to}$ Pred error is similar to the Pred → GT error.

\begin{table*}[!ht]
	\centering
	\caption{The completion errors (Pred $\mathrm{\to }$ GT/GT $\mathrm{\to }$ Pred) of the incomplete vehicle point clouds in autonomous driving}
	\label{tab2}
	\begin{tabular}{m{1.1in} m{1in} m{0.6in} m{0.6in} m{0.6in} m{0.6in}} \hline\noalign{\smallskip}
		\textbf{Number of Points} & \textbf{Algorithm} & \textbf{Vehicle 1} & \textbf{Vehicle 2} & \textbf{Vehicle 3} & \textbf{Mean} \\ \noalign{\smallskip}\hline\noalign{\smallskip}
	\multirow{2}*{2048}	 & Original PF-Net & 1.030/0.995 & 1.300/1.505 & 1.292/1.220 & 1.207/1.240 \\  
		& Improved PF-Net & 1.092/0.985 & 1.270/1.354 & 1.322/1.200 & 1.228/1.180 \\ \noalign{\smallskip}\noalign{\smallskip}
	\multirow{2}*{4048}	 & Original PF-Net & 0.744/0.937 & 0.937/1.600 & 0.910/1.223 & 0.864/1.253 \\ 
		& Improved PF-Net & 0.775/0.922 & 0.971/1.435 & 0.898/1.236 & 0.881/1.198 \\  \noalign{\smallskip}\noalign{\smallskip}
	\multirow{2}*{6048}	 & Original PF-Net & 0.627/0.934 & 0.804/1.571 & 0.738/1.222 & 0.723/1.242 \\  
		& Improved PF-Net & 0.659/0.924 & 0.820/1.453 & 0.760/1.222 & 0.746/1.200 \\ \noalign{\smallskip}\noalign{\smallskip}
		\multirow{2}*{8048} & Original PF-Net & 0.582/0.936 & 0.785/1.539 & 0.704/1.247 & 0.690/1.240 \\  
		& Improved PF-Net & 0.641/0.934 & 0.753/1.536 & 0.704/1.244 & 0.699/1.238 \\ \noalign{\smallskip}\noalign{\smallskip}
		\multirow{2}*{10048} & Original PF-Net & 0.584/0.945 & 0.682/1.603 & 0.653/1.218 & 0.640/1.255 \\  
		& Improved PF-Net & 0.620/0.952 & 0.694/1.502 & 0.651/1.219 & 0.655/1.224 \\ \noalign{\smallskip}\hline 
	\end{tabular}
	
\end{table*}

\begin{figure*}[!h]
	\centering
	\includegraphics[width=\textwidth]{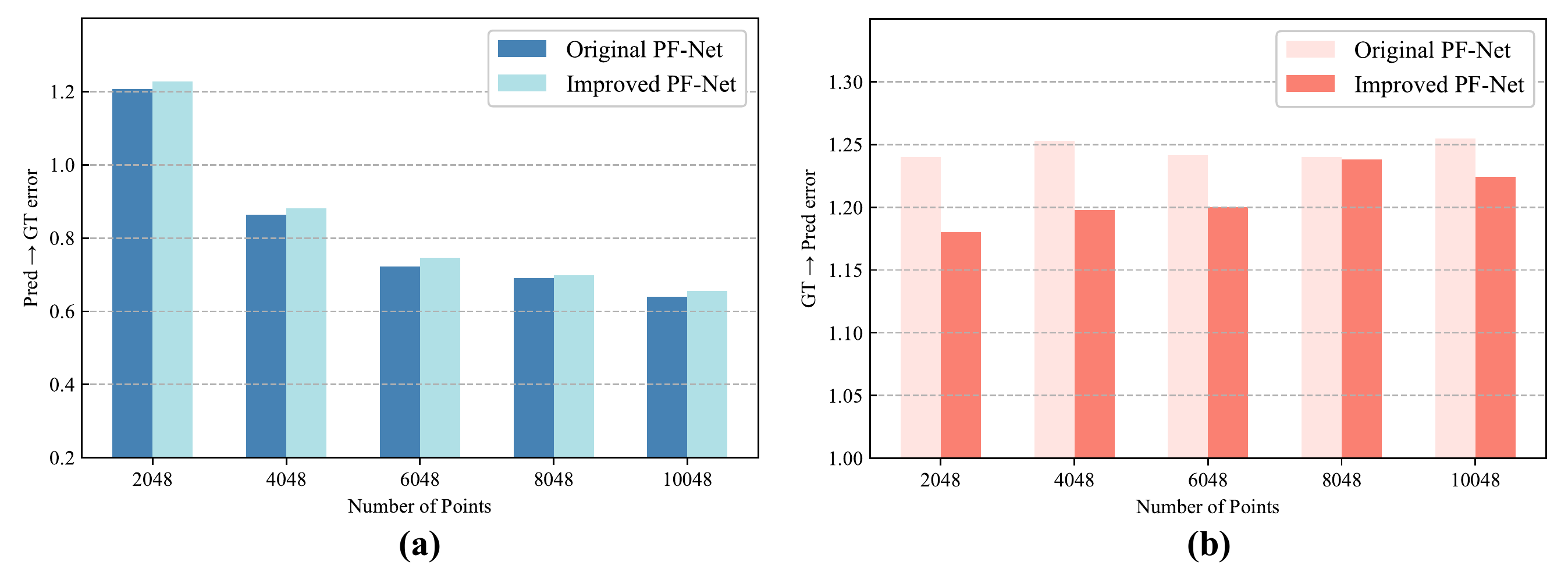}
	\caption{Average Pred $\mathrm{\to}$ GT error and GT $\mathrm{\to}$ Pred error of point cloud completion for different sizes ((a) Average Pred $\mathrm{\to}$ GT error; (b) Average GT →$\mathrm{\to}$ Pred error.)}
	\label{fig6}       % Give a unique label
\end{figure*}

As listed in Table \ref{tab2}, the point cloud repair for Vehicle 1 has the best performance, and the accuracy of the point cloud completion for Vehicle 2 is the lowest. Moreover, when completing the same vehicle, as the point cloud size (number of points) increases, the Pred $\mathrm{\to}$ GT error of completion tends to decrease significantly, and the GT $\mathrm{\to}$ Pred error of completion doesn't decrease or increase significantly (see Fig. \ref{fig6}). 

In conclusion, as shown in Fig. \ref{fig6}, The average Pred $\mathrm{\to}$ GT error of the improved PF-Net is slightly larger (approximately 2.1\%) than that of the original PF-Net, and the average GT $\mathrm{\to}$ Pred error of the improved PF-Net is slightly smaller (approximately 3.1\%) than that of the original PF-Net. The results indicated that there is no significant difference in the accuracy of the improved PF-Net and the original PF-Net, and both increase with the increase of the number of points.

\subsubsection{Completion Efficiency on Vehicle Point Clouds of Different Scales}

Since the point cloud completion serves self-driving vehicles, we focus on the inference efficiency of deep neural networks in this paper. The inference process of completion networks consists of two modules: (1) sampling module and (2) generation module. As shown in Fig. \ref{fig2}, the sampling module includes two downsampling processes. First, the input vehicle point cloud needs to be downsampled into a point cloud with 1024 points. Second, the point cloud 1024 points are downsampled into a point cloud with 512 points. Finally, we can obtain three scales point clouds that meet the input requirements of the generation module (i.e., the generator).
\begin{table*}[!ht]
	\centering
	\caption{Running time (ms) of point cloud completion on vehicle point clouds of different scales (Where $t_1$ is the sampling module running time, $t_2$ is the generation module running time, and $t_3$ is the total time, i.e., $t_3=t_1+t_2$.)}
	\label{tab3}
\begin{tabular}{m{1.1in}m{1.2in}m{0.3in}m{0.3in}m{0.3in}m{0.3in}m{0.3in}m{0.3in}m{0.3in}m{0.3in}m{0.3in}} \hline\noalign{\smallskip}
	\multirow{2}*{\textbf{Number of Points}} & \multirow{2}*{\textbf{Algorithm}} & \multicolumn{3}{p{0.9in}<{\centering}}{\textbf{Vehicle 1}} & \multicolumn{3}{p{0.9in}<{\centering}}{\textbf{Vehicle 2}} & \multicolumn{3}{p{0.9in}<{\centering}}{\textbf{Vehicle 3}} \\  
	\textbf{} & \textbf{} & \textbf{t1} & \textbf{t2} & \textbf{t3} & \textbf{t1} & \textbf{t2} & \textbf{t3} & \textbf{t1} & \textbf{t2} & \textbf{t3} \\ \noalign{\smallskip}\hline\noalign{\smallskip}
	\multirow{2}*{2048} & Original PF-Net & 271.1 & 5.9 & \textbf{277.0} & 263.3 & 6.0 & \textbf{269.3} & 279.2 & 5.0 & \textbf{284.2} \\  
	& Improved PF-Net & 6.0 & 7.0 & \textbf{13.0} & 14.9 & 5.0 & \textbf{19.9} & 6.0 & 5.0 & \textbf{11.0} \\ \noalign{\smallskip}\noalign{\smallskip} 
	\multirow{2}*{4048}  & Original PF-Net & 392.9 & 6.9 & \textbf{399.8} & 396.9 & 5.0 & \textbf{401.9} & 411.0 & 5.9 & \textbf{416.9} \\  
	& Improved PF-Net & 7.9 & 6.0 & \textbf{13.9} & 9.0 & 6.0 & \textbf{15.0} & 8.9 & 5.9 & \textbf{14.8} \\ \noalign{\smallskip}\noalign{\smallskip} 
	\multirow{2}*{6048}  & Original PF-Net & 527.6 & 6.0 & \textbf{533.6} & 533.7 & 6.0 & \textbf{539.7} & 524.6 & 6.0 & \textbf{530.6} \\ 
	& Improved PF-Net & 11.1 & 6.0 & \textbf{17.1} & 15.0 & 6.0 & \textbf{21.0} & 11.0 & 6.0 & \textbf{17.0} \\ \noalign{\smallskip}\noalign{\smallskip} 
	\multirow{2}*{8048}   & Original PF-Net & 637.3 & 6.0 & \textbf{643.3} & 652.2 & 6.0 & \textbf{658.2} & 639.8 & 6.0 & \textbf{645.8} \\ 
	& Improved PF-Net & 14.0 & 5.1 & \textbf{19.1} & 15.9 & 5.0 & \textbf{20.9} & 11.9 & 6.0 & \textbf{17.9} \\ \noalign{\smallskip}\noalign{\smallskip} 
	\multirow{2}*{10048} & Original PF-Net & 759.9 & 5.3 & \textbf{765.2} & 775.9 & 6.0 & \textbf{781.9} & 790.9 & 6.8 & \textbf{797.7} \\ 
	& Improved PF-Net & 14.9 & 5.0 & \textbf{19.9} & 17.9 & 6.0 & \textbf{23.9} & 14.0 & 6.0 & \textbf{20.0} \\ \noalign{\smallskip}\hline
\end{tabular}
\end{table*}

\begin{figure*}[!h]
	\centering
	\includegraphics[width=\textwidth]{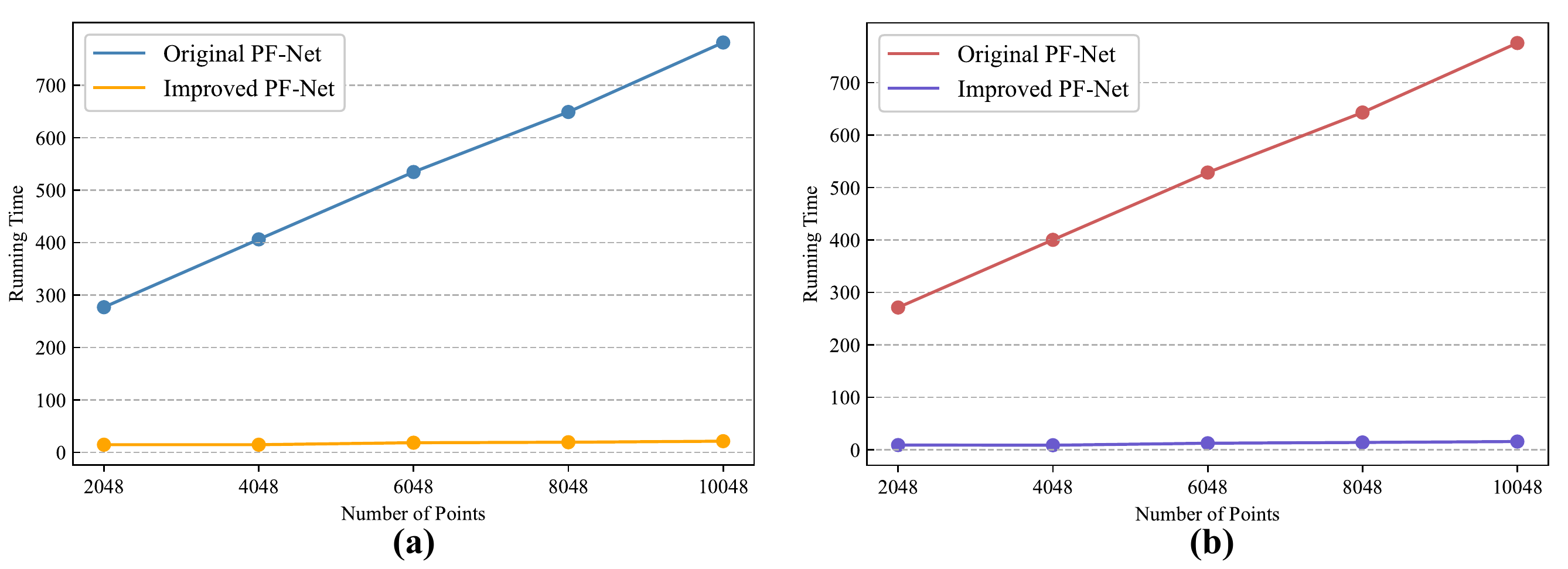}
	\caption{Average total time and average sampling module running time in original PF-Net and improved PF-Net ((a) Average total time; (b) Average sampling module running time.)}
	\label{fig7}       % Give a unique label
\end{figure*}

As listed in Table \ref{tab3}, when the number of points and the model are the same, there is little difference in the running time of each module for the completion of different vehicle point clouds. As the number of points increases, the sampling module running time and the total time tend to increase when the model is the same, and the sampling module running time and the total time of the original PF-Net are much larger than those of the improved PF-Net when the number of points is the same (see Fig. \ref{fig7}). In any case, the generation module running time is always approximately 6 ms. 

\begin{figure*}[!h]
	\centering
	\includegraphics[width=\textwidth]{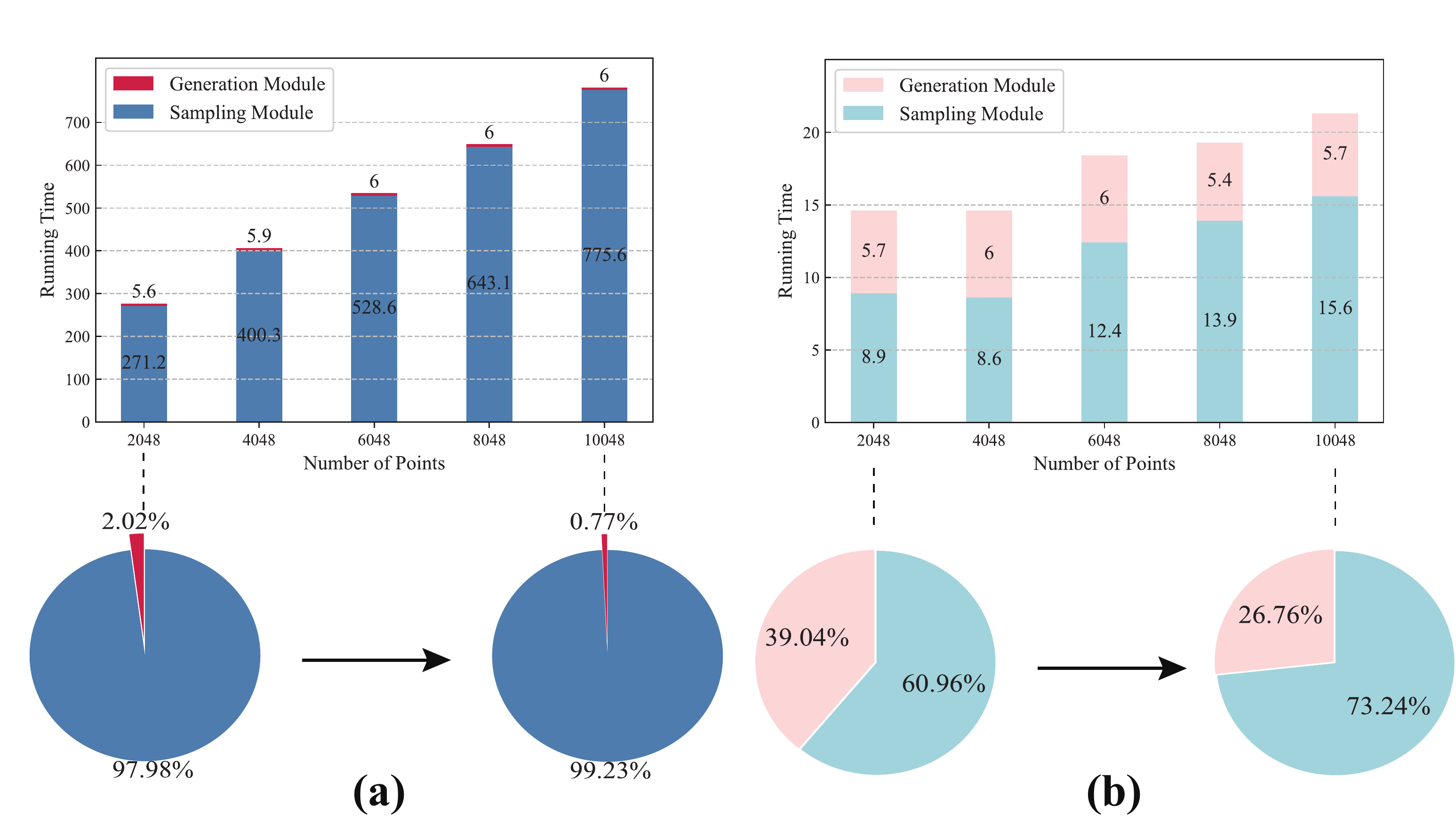}
	\caption{Average running time (ms) bar plots of sampling module and generation module ((a) Original PF-Net; (b) Improved PF-Net.)}
	\label{fig8}       % Give a unique label
\end{figure*}

As shown in Fig. \ref{fig8}(a), for the original PF-Net, the sampling module running time is much larger (265.6$ \sim $\\769.6 ms) than the generation module running time, and the difference between them increases as the number of points increases (the percentage of the sampling module running time in the total time increases from 97.98\% to 99.23\%), which means that the sampling module is the key factor in the efficiency of point cloud completion, and its impact increases with the number of points. As shown in Fig. \ref{fig8}(b), for the improved PF-Net, the sampling module running time is only a little larger (2.2$ \sim $9.9 ms) than the generation module running time, and the difference between them increases as the number of points increases (the percentage of the sampling module running time in the total time increases from 60.96\% to 73.24\%).

\begin{figure}[!h]
	\centering
	\includegraphics[width=0.5\textwidth]{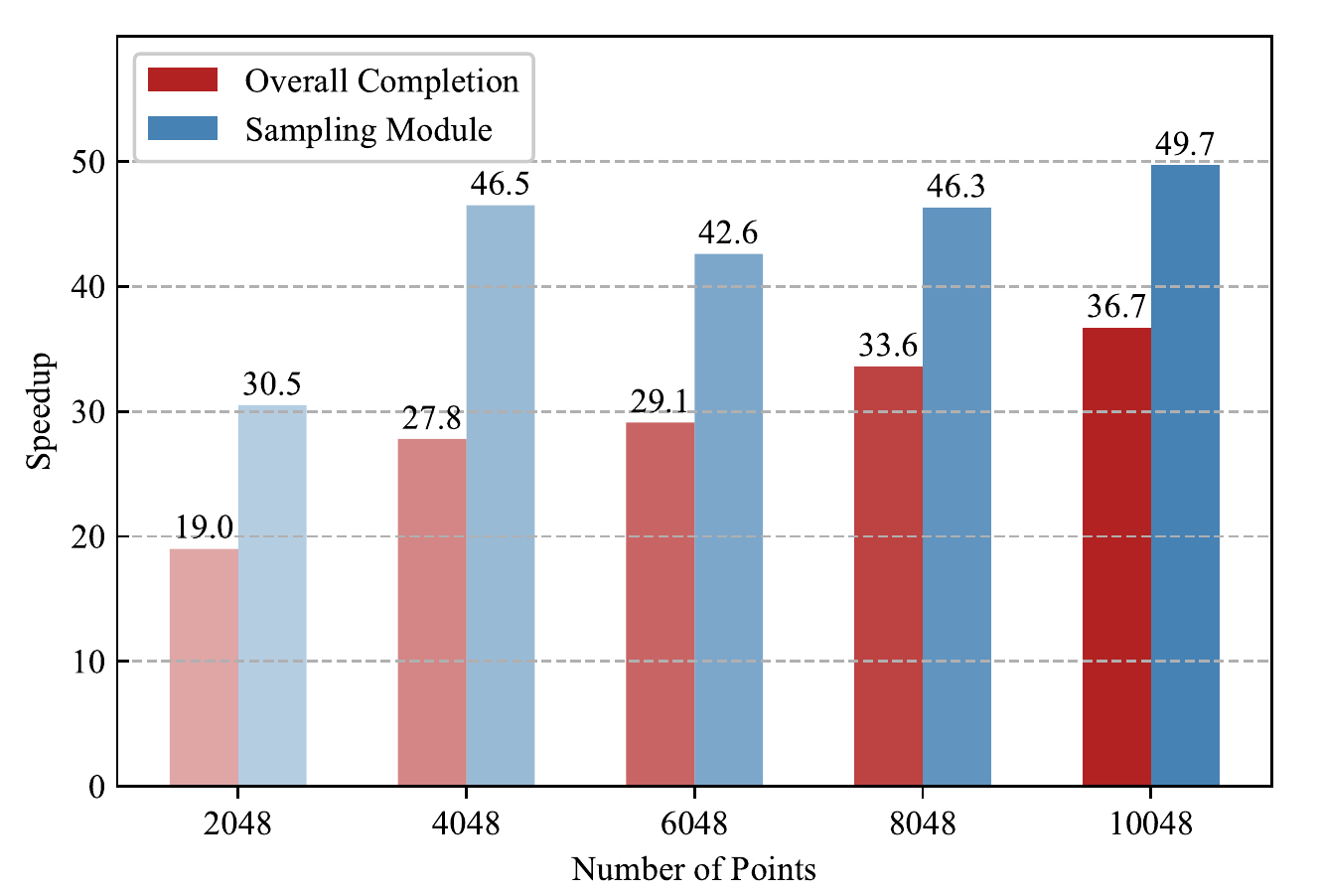}
	\caption{Speedups of overall completion and sampling module in the improved PF-Net ($speedup\mathrm{=}\frac{t_P}{t_{IP}}$, where $t_P$ is the running time of overall completion or sampling module in the original PF-Net, and $t_{IP}$ is the running time of overall completion or sampling module in the improved PF-Net.)}
	\label{fig9}       % Give a unique label
\end{figure}

As shown in Fig. \ref{fig9}, compared with the original PF-Net, the efficiency of the proposed completion networks (the improved PF-Net) is greatly improved, and the speedup of overall completion is 19$ \sim $36.7. Moreover, compared with the IFPS algorithm, the efficiency of the proposed Cell-IFPS is greatly improved, the speedup of the sampling module using the proposed Cell-IFPS is 30.5$ \sim $49.7.

The aforementioned comparative results indicate that:

(1)	The efficiency of the sampling module is the key factor in the efficiency of point cloud completion, and its impact increases with the number of points.

(2)	Compared with the original PF-Net, the efficiency of the improved PF-Net is greatly improved due to the efficient Cell-IFPS, and the maximum speedup is 36.7.

(3)	Compared with the IFPS algorithm, as the scale of the point clouds increases, the acceleration effect of the proposed Cell-IFPS increases, and the maximum speedup is 49.7.

\subsection{Completion Results on a Real-world Vehicle Point Cloud}

In this section, for the incomplete vehicle point cloud from the real self-driving process, we specifically introduce the completion efficiency of the original PF-Net and the improved PF-Net, and briefly describe the completion effect.

\begin{figure}[!h]
	\centering
	\includegraphics[width=0.5\textwidth]{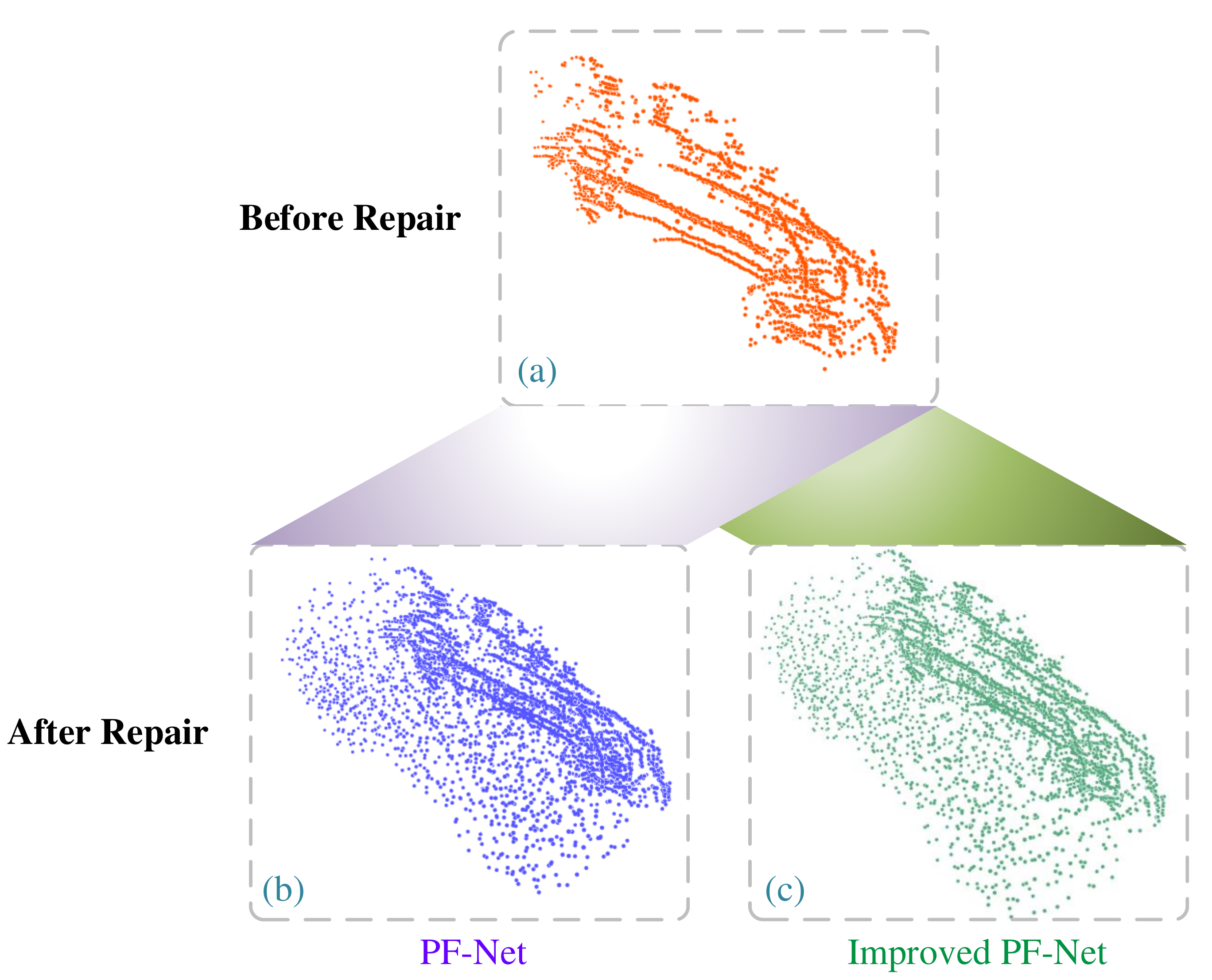}
	\caption{An Illustration of vehicle point cloud completion on a real-world vehicle point cloud in self-driving}
	\label{fig10}       % Give a unique label
\end{figure}

As shown in Fig. \ref{fig10}, for the incomplete vehicle point cloud, both original PF-Net and improved PF-Net successfully complete the missing area of the vehicle so that the point cloud has the complete shape of the vehicle after the repair, and the missing point clouds generated by the original PF-Net and the improved PF-Net are almost identical.

As listed in Table \ref{tab4}, the total running time of the improved PF-Net is much less than that of the original PF-Net, and the speedup is 30.9; the sampling module running time of the improved PF-Net is much less than that of the original PF-Net, and the speedup is 21.7. The improved PF-Net performs a point cloud completion in only 20 ms during autonomous driving, i.e., the speed of the improved PF-Net is approximately 46 fps during autonomous driving, which satisfies the requirement for real-time vehicle point cloud completion for autonomous driving.
\begin{table}[!ht]
	\centering
	\caption{Running time (ms) of point cloud completion on a real self-driving vehicle point cloud ($speedup\mathrm{=}\frac{t_P}{t_{IP}}$, where $t_P$ is the total running time or the sampling module running time in the original PF-Net, and $t_{IP}$ is the total running time or the sampling module running time in the improved PF-Net.)}
	\label{tab4}
\begin{tabular}{m{0.7in} m{0.7in} m{0.7in} m{0.5in}} \hline\noalign{\smallskip}
	\textbf{Module} & \textbf{Original PF-Net} & \textbf{Improved PF-Net} & \textbf{Speedup} \\ 
	\noalign{\smallskip}\hline\noalign{\smallskip}
	Sampling Module & 368.1 & 11.9 & 30.9 \\  \noalign{\smallskip}\noalign{\smallskip} 
	Generation Module & 4.9 & 5.3 & 0.9 \\ \noalign{\smallskip}\noalign{\smallskip} 
	Total & 373.0 & 17.2 & 21.7 \\ 
	\noalign{\smallskip}\hline
\end{tabular}

\end{table}

\section{Discussion}
\label{dis}
\subsection{Advantages of the proposed deep learning method}
The advantage of the proposed method is to consider the efficiency of point cloud completion networks for self-driving vehicles. More specifically, in this paper, we found that the sampling algorithm is the key factor that restricts the efficiency of point cloud completion networks. Therefore, we proposed an efficient sampling algorithm Cell-IFPS by combining the one-time sampling algorithm Cell sampling with the incremental sampling algorithm IFPS, which can precisely control the number of sampling points. The proposed Cell-IFPS substantially improves the inference speed of the original PF-Net without using a parallel strategy for acceleration. In autonomous driving scenarios, the tasks are generally time-critical. As listed in Table \ref{tab4}, the point cloud completion efficiency of the original PF-Net is difficult to meet the needs of self-driving, only 3 fps, and the point cloud completion efficiency of the improved PF-Net is 46 fps, which is enough to meet the requirement of speed. Moreover, as shown in Fig. \ref{fig7}, the efficiency of the improved PF-Net is insensitive to changes in the size of the point cloud, i.e., an increase in the number of points doesn’t lead to a sharp decrease in its efficiency, which gives it the potential to serve tasks such as the 3D reconstruction of large-scale point clouds.

On the other hand, for the incomplete vehicle point cloud repair, the accuracy of improved PF-Net is almost the same as that of the original PF-Net, which performs well in accuracy and surpassing the widely used point cloud generation model PCN [3,10]. The improved PF-Net with high accuracy can be effectively used to realize the real-time vehicle point cloud completion to enhance the safety of autonomous driving.

\subsection{Shortcomings of the proposed deep learning method}

The shortcoming of the proposed method is that the accuracy of the point cloud completion may be reduced or even the point cloud completion may fail when the scale of the input point cloud is too small. According to the network architecture, the input point cloud needs to use downsampling to extract the feature points, in this paper, the input point cloud is downsampled by 1024 points and 512 points, which means that the improved PF-Net cannot complete the point cloud when the size of the input point cloud is less than 1024 points. On the other hand, for the real self-driving data, we lack the real point cloud of the missing part of the incomplete vehicle point cloud. Therefore, we don’t use the corresponding evaluation metric to evaluate the completion accuracy.

\subsection{Outlook and Future Work}
In the future, we plan to apply the improved PF-Net to repair incomplete point clouds that include more kinds of obstacles (e.g., trees, roadblocks, telegraph poles) in autonomous driving. Further, the proposed efficient uniform sampling method, Cell-IFPS, will be applied to more tasks for point cloud (e.g., 3D reconstruction, object recognition, point cloud segmentation) due to the sampling being one of the foundations of point cloud processing.

There are several techniques to improve the efficiency of deep learning, including CPU/GPU-accelerated parallel technology and TensorRT technology \cite{18,26,27}. In the future, we will use parallel technology to speed up the sampling process or use the TensorRT technology to accelerate the generation process at the cost of slightly reducing the accuracy.

Moreover, the deployment of deep learning in actual production is still difficult. Compared with Python language, the deployment of deep learning algorithms developed by C/C++ language is more convenient. Therefore, we will consider developing or using API to convert the Python version of the improved PF-Net to the C/C++ version to facilitate deployment.

\section{Conclusion}
\label{con}
In this paper, to meet the requirements of algorithm efficiency for incomplete point cloud completion in the process of self-driving, we proposed the improved PF-Net to repair incomplete vehicle point cloud accurately and efficiently in autonomous driving. The essential idea of the method is to use efficient deep learning for real-time incomplete vehicle point cloud completion to improve the safety of self-driving vehicles. In the proposed method, an efficient down sampling combining incremental sampling and one-time sampling, Cell-IFPS, is presented to significantly improve the inference speed of the original PF-Net. To evaluate the performance of the proposed method, a real dataset is used, and an autonomous driving scene is created, where three incomplete vehicle point clouds with 5 different sizes are set for three autonomous driving situations. The results show that: (1) inefficient sampling module is the key to restrict the efficiency of the original PF-Net; (2) the proposed Cell-IFPS can greatly improve the efficiency of the original PF-Net without using a parallel strategy for acceleration, and the enhancement effect increases with the increase of point cloud scale; (3) the improved PF-Net has far greater speed than the original PF-Net and almost the same accuracy as the original PF-Net, which is the state-of-the-art point cloud repair network in accuracy. The improved PF-Net has the capability to complete incomplete vehicle point clouds of self-driving vehicles.

\begin{acknowledgements}
	This research was jointly supported by the National Natural Science Foundation of China (Grant No. 11602235), and the Fundamental Research Funds for China Central Universities (2652018091).
\end{acknowledgements}




%\begin{acknowledgements}
%\end{acknowledgements}


\bibliographystyle{unsrt}
\bibliography{Ref}   % name your BibTeX data base

%\begin{thebibliography}{}

\end{document}